\documentclass[conference]{IEEEtran}
\IEEEoverridecommandlockouts

\usepackage[colorlinks,urlcolor=blue,linkcolor=blue,citecolor=blue]{hyperref}

\usepackage{color,array}
\usepackage{cite}
\usepackage{amsmath,amssymb,amsfonts}
\usepackage{algorithm}
\usepackage{algorithmic}
\usepackage{graphicx}
\usepackage{textcomp}
\usepackage{xcolor}
\usepackage{tabularx}
\usepackage{multirow}
\usepackage{adjustbox}
\usepackage[edges]{forest}
\usepackage{array}
\usepackage{rotating}
\usepackage{float}
\usepackage{multirow}

\usepackage{tikz}
\usetikzlibrary{arrows.meta, positioning}
\begin{document}

\title{A Survey of Sustainability in Large Language Models: Applications, Economics, and Challenges}

\author{
\IEEEauthorblockN{Aditi Singh\textsuperscript{1}, Nirmal Prakashbhai Patel\textsuperscript{1}, Abul Ehtesham\textsuperscript{2}, Saket Kumar\textsuperscript{3}, Tala Talaei Khoei\textsuperscript{4}}
\IEEEauthorblockA{
\textsuperscript{1}\textit{Department of Computer Science, Cleveland State University, USA} \\
\textsuperscript{2}\textit{The Davey Tree Expert Company, USA} \\
\textsuperscript{3}\textit{The Mathworks, USA} \\
\textsuperscript{4}\textit{Khoury College of Computer Science, 
Roux Institute at Northeastern University,  USA} \\
a.singh22@csuohio.edu, n.patel27@vikes.csuohio.edu, abul.ehtesham@davey.com,  \\ 
saketk@mathworks.com, t.talaeikhoei@northeastern.edu
}
}

\maketitle
\begin{abstract}
Large Language Models (LLMs) have transformed numerous domains by providing advanced capabilities in natural language understanding, generation, and reasoning. Despite their groundbreaking applications across industries such as research, healthcare, and creative media, their rapid adoption raises critical concerns regarding sustainability. This survey paper comprehensively examines the environmental, economic, and computational challenges associated with LLMs, focusing on energy consumption, carbon emissions, and resource utilization in data centers. By synthesizing insights from existing literature, this work explores strategies such as resource-efficient training, sustainable deployment practices, and lifecycle assessments to mitigate the environmental impacts of LLMs. Key areas of emphasis include energy optimization, renewable energy integration, and balancing performance with sustainability. The findings aim to guide researchers, practitioners, and policymakers in developing actionable strategies for sustainable AI systems, fostering a responsible and environmentally conscious future for artificial intelligence.
\end{abstract}

\begin{IEEEkeywords}
Large Language Models (LLMs), Sustainability in AI, AI Economics and Cost Analysis, Energy Efficiency in AI, Environmental Impact of AI, carbon footprint.
\end{IEEEkeywords}

\section{Introduction}
The rapid advancement of large language models (LLMs) has generated significant interest and investment across various sectors of artificial intelligence. With their impressive capabilities, LLMs have the potential to transform everyday experiences, facilitating more intuitive interactions in applications ranging from customer support to content creation. However, as this technological boom unfolds, it is crucial to prioritize sustainability and the societal benefits that can arise from these advancements.

Sustainability in the context of LLMs involves a thorough evaluation of their environmental, economic, and social impacts. The development and deployment of these models require substantial computational resources, which raises concerns about energy consumption and carbon emissions. Addressing these challenges is vital to ensuring that the benefits of LLMs do not come at the expense of planetary health. Exploring sustainable practices in LLM development can maximize their positive impact while minimizing their ecological footprint.

This paper highlights the transformative potential of LLMs while emphasizing the importance of sustainable practices that benefit both society and the environment. By examining the economic and social implications of LLM development and deployment, it aims to provide insights into how these technologies can be harnessed responsibly to improve the quality of life for individuals and communities. Several research questions are formulated to navigate the complexities of LLMs and their role in fostering a sustainable future.

The following sections delve into these research questions and inquiries to provide a comprehensive analysis of the sustainability of LLMs in today’s rapidly evolving technological landscape.

\subsection{Research Questions and Inquiries}
\begin{figure}[htbp]
\centering
\begin{forest}
for tree={%
                forked edges,
                grow'=0,
                draw,
                rounded corners,
                node options={align=center,},
                text width=2.5cm, % Adjusts node width
                level distance=1.5cm, % Vertical spacing
                sibling distance=1.8cm, % Horizontal spacing
            },
[Research Questions
    [Tasks and Domains of LLMs, fill=blue!20
        [Identify Practical Applications, fill=blue!10]
    ]
    [Environmental and Economic Costs, fill=green!20
        [Evaluate Training \\ and Deployment Costs, fill=green!10]
        [Assess Sustainability, fill=green!10]
    ]
    [Sustainable and Energy-Efficient LLMs, fill=yellow!20
        [Explore Energy \\ Reduction Strategies, fill=yellow!10]
    ]
]
\end{forest}
\caption{Hierarchical Tree of Research Questions and Objectives for LLM Sustainability}
\label{fig:research_questions}
\end{figure}
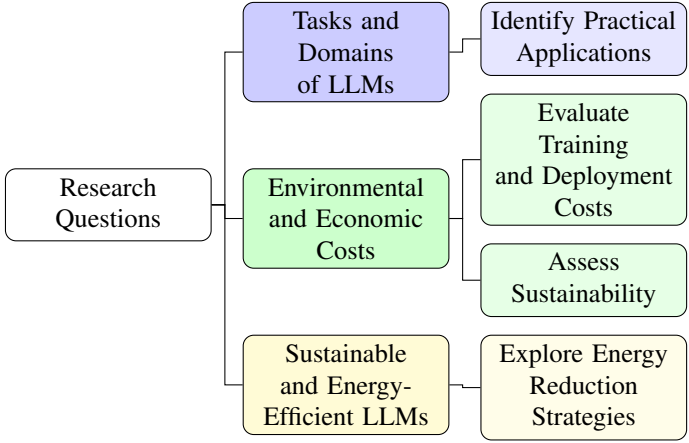

The research questions explored in this paper (Figure \ref{fig:research_questions}), along with their corresponding inquiries, are as follows:

\begin{itemize}
    \item \textbf{Research Question 1: What are the specific tasks and domains where LLMs excel} 
        \begin{itemize}
            \item Objective: Identify the practical LLM applications and their potential.
        \end{itemize}
    \item \textbf{Research Question 2: What are the environmental and economic costs associated with the development and deployment of LLMs?} 
        \begin{itemize}
            \item Objective: To evaluate the impact of training and operational costs on sustainability and resource allocation.
        \end{itemize}
    \item \textbf{Research Question 3: How can LLMs evolve to become more sustainable and energy-efficient?} 
        \begin{itemize}
            \item Objective: Explore strategies and technologies that reduce energy consumption and mitigate environmental impact.
        \end{itemize}
\end{itemize}
By addressing these objectives, this paper aims to provide actionable insights into the sustainability challenges of LLMs and offer a comprehensive framework for their responsible development and use.

The paper is structured as follows: Section 2 surveys LLMs applications. Section 3 examines the environmental and economic costs associated with LLM development and deployment. Section 4 discusses sustainable practices and strategies for resource optimization, and Section 5 concludes with future directions and actionable recommendations for sustainable AI development.

\section{LLM Applications}

At present, large language models (LLMs) are being applied across a wide array of industries \cite{sin23}, showcasing their versatility and transformative potential. The applications of LLMs can be categorized into several key areas, as summarized in Table \ref{table:llm-applications}. 
\begin{table*}[ht!]
    \caption{Different Types of LLM Applications}
    \label{table:llm-applications}
    \centering
    \fontsize{8}{9}\selectfont
    \renewcommand{\arraystretch}{1.2} % Adjust row spacing
    \begin{tabularx}{\textwidth}{|c|X|X|}
    % {|p{0.15\textwidth}|p{0.35\textwidth}|p{0.35\textwidth}|}
    \hline
    \textbf{Type of Use} & \textbf{Description} & \textbf{Use Case in Industry} \\
    \hline
    Text-to-Text & Generates text from text descriptions &  Education, Marketing, e-commerce. \\
    \hline
    Text-to-Image & Generate images from text descriptions. & Marketing (ad visuals), e-commerce (product previews). \\
    \hline
    Text-to-Video & Create videos or animations from text inputs. & Film industry (storyboarding), education (video lessons). \\
    \hline
    Text-to-Audio & Convert text into speech or environmental sounds. & Virtual assistants, audiobooks, accessibility tools. \\
    \hline
    Audio-to-Text & Convert audio or speech into text. & Transcription services, customer support analysis. \\
    \hline
    Text-to-3D & Create 3D models or objects from text inputs. & Gaming (3D assets), architecture (concept modeling). \\
    \hline
    Image-to-3D & Generate 3D objects or avatars from photos. & Fashion (virtual try-ons), gaming (avatar creation). \\
    \hline
    Video-to-3D & Build 3D environments from video inputs. & VR (immersive experiences), film production (3D scenes). \\
    \hline
    \end{tabularx}
\end{table*}

\begin{itemize}
    \item \textbf{Text-to-Text}: Text-to-Text Large Language Models (LLMs) enable the generation of text from textual descriptions, with applications such as text summarization\cite{zhang2024systematicsurveytextsummarization, fang2024multillmtextsummarization, basyal2023textsummarizationusinglarge, liu2024learningsummarizelargelanguage}, multilingual text generation\cite{uthus2023mlongt5multilingualefficienttexttotext, hu2024gentranslatelargelanguagemodels, srivastava2025lolaopensourcemassively, wei2023polylmopensourcepolyglot}, and question answering \cite{singhal2023expertlevelmedicalquestionanswering, sun2024harnessingmultirolecapabilitieslarge, kamalloo-etal-2023-evaluating} using RAG\cite{alawwad2024enhancingtextbookquestionanswering}. These models are widely used for condensing information, enabling cross-lingual communication, and developing intelligent systems like chatbots and virtual assistants. Their ability to process and generate coherent, contextually appropriate text across domains demonstrates their transformative potential in automating complex language tasks.
    \item \textbf{Text-to-Image}: This application enables the generation of images from text descriptions, making it particularly valuable in marketing and e-commerce. Companies can create visually appealing advertisements and product previews, enhancing customer engagement and driving sales \cite{zhang2023text, surveysing2023, wan2024survey}.
    
    \item \textbf{Text-to-Video}: By transforming text-based inputs into videos or animations, this application has significant implications for the film industry, where it can be used for storyboarding and pre-visualization. Additionally, educational institutions can utilize this technology to create engaging video lessons, facilitating better learning experiences. \cite{surveysing2023, zhang2023text, cho2024sora, sun2024sora, xing2023survey}.
    
    \item \textbf{Text-to-Audio}: The ability to convert text into human-like speech or environmental sounds is revolutionizing virtual assistants and accessibility tools. This application allows for more natural interactions with technology, helping to improve user experience and accessibility for individuals with disabilities. \cite{majumder2024tango2aligningdiffusionbased, liu2023audioldm, huang2023makeanaudio,huang2023makeanaudio2,liu2023baton, hai2024ezaudio, liu2024audiolcm, dong2023clipsonic, yuan2023realaudioldm, ghosal2023tango}
    
    \item \textbf{Image-to-Video}: This application allows for generating video based on the given image \cite{liu2024physgen, ni2023latentflow, guo2023i2vadapter, xu2024camco}. 
    
    \item \textbf{Text-to-3D}: Generating 3D models or objects from text inputs is a game-changer for industries such as gaming and architecture. This capability allows for rapid prototyping and the creation of immersive experiences in virtual environments. \cite{jain2023zero, lee2024textto3dshapegeneration, ding2023textto3d, shriram2024realmdreamertextdriven3dscene, raj2023dreambooth3dsubjectdriventextto3dgeneration, chen2023it3dimprovedtextto3dgeneration}
    
    \item \textbf{Image-to-3D}: By creating 3D objects or avatars from photos, this application finds relevance in fashion for virtual try-ons and in gaming for avatar creation, enhancing user engagement and personalization. \cite{hong20233dllminjecting3dworld}
    
    \item \textbf{Video-to-3D}: The ability to build 3D environments from video inputs opens up exciting possibilities in virtual reality (VR) and film production, allowing for more immersive storytelling and experiences.
\end{itemize}

Overall, the diverse applications of LLMs demonstrate their potential to drive innovation and efficiency across various industries. As these technologies continue to evolve, their impact on content creation, interaction, and understanding is expected to grow.

\section{Cost of LLM Development}

The development of large language models (LLMs) entails significant financial investments, primarily due to the extensive computational resources required for training and deployment. This subsection will explore the various costs associated with LLM development, including:

\begin{itemize}
    \item \textbf{Training Costs}: The computational expenses incurred during the training phase, which involve powerful hardware and large datasets.
    
    \item \textbf{Resource Allocation}: The costs associated with data storage, model maintenance, and the expertise required for managing these sophisticated systems.
    
    \item \textbf{Deployment Costs}: The expenses related to integrating LLMs into existing systems, including infrastructure and operational costs.
\end{itemize}

Understanding these costs Table \ref{tab:llm_costs} is crucial for assessing the overall feasibility and sustainability of LLMs, especially when considering their long-term impact on various industries.
\begin{table*}[h!]
\caption{Cost Components for Training, Maintaining, and Deploying Large Language Models (LLMs)}
\label{tab:llm_costs}
\centering
\fontsize{8}{10}\selectfont
\renewcommand{\arraystretch}{1.3} % Adjust row spacing
\begin{tabular}{|p{3cm}|p{3cm}|p{8cm}|p{3cm}|}
\hline
\textbf{Cost Type} & \textbf{Cost Components} & \textbf{Description} & \textbf{Estimated Cost (USD)} \\ \hline

\multirow{2}{3cm}{\textbf{Training LLM}} & Energy & Power for GPU clusters, consumption depends on model size and training duration (e.g., GPT-3 training required 1 GWh/day). & \$500,000 - \$5 million per training run \cite{poole_what_2024} \\ \cline{2-4}
 & Computing & Use of specialized hardware like NVIDIA A100/H100 GPUs or TPUs. For PaLM, training took 8,404,992 TPUv4-core hours. & \$9M - \$23M per training cycle \cite{chowdhery_palm_2022}, \cite{noauthor_estimating_2022} \\ \hline

\multirow{2}{3cm}{\textbf{Model Development}} & Data Collection and Cleaning & Acquisition, filtering, and preparation of large, high-quality datasets. & \$50,000 - \$100,000 \cite{samarpit_what_2023}, \cite{team_understanding_2024} \\ \cline{2-4}
 & Model Fine-Tuning & Additional compute and expert involvement required to tailor models for specific use cases. & \$50,000 - \$500,000 \cite{noauthor_how_nodate}, \cite{samarpit_what_2023} \\ \hline

\multirow{4}{3cm}{\textbf{Maintain Server}} & Storage & Storing model weights, checkpoints, and datasets (cloud storage costs vary). & \$10,000 - \$100,000/month \cite{team_understanding_2024}, \cite{noauthor_how_nodate} \\ \cline{2-4}
 & Electric & Continuous power for servers and cooling systems (data centers consume ~1 GWh/day for models like ChatGPT). & \$5,000 - \$50,000/month \cite{noauthor_what_nodate}, \cite{poole_what_2024} \\ \cline{2-4}
 & Energy Resource & Usage of renewable energy to reduce environmental impact (scales with model size). & Initial setup \$50,000+; \$2,000 - \$20,000/month \cite{chowdhery_palm_2022}, \cite{samarpit_what_2023} \\ \cline{2-4}
 & Cooling Systems & Data centers with high-performance GPUs require extensive cooling infrastructure. & \$10,000 - \$100,000/month \cite{samarpit_what_2023}, \cite{poole_what_2024} \\ \hline

\textbf{Software Licensing} & Proprietary Software/Frameworks & Licensing fees for tools like TensorFlow, PyTorch, etc. & \$5,000 - \$50,000/year \cite{chowdhery_palm_2022}, \cite{poole_what_2024} \\ \hline

\multirow{2}{3cm}{\textbf{Compliance \& Security}} & Data Security and Compliance & Adherence to GDPR, HIPAA, and other regulations during training and deployment. & \$20,000 - \$100,000/year \cite{samarpit_what_2023}, \cite{team_understanding_2024} \\ \cline{2-4}
 & AI Ethics and Bias Audits & Ensuring fairness and ethical compliance through regular audits. & \$50,000 - \$150,000 per audit \cite{team_understanding_2024}, \cite{samarpit_what_2023} \\ \hline

\textbf{Personnel} & Skilled Workforce & Salaries for AI researchers, engineers, and cloud DevOps experts. & \$500,000 - \$1 million/year for a small team \cite{team_understanding_2024}, \cite{noauthor_what_nodate} \\ \hline

\multirow{2}{3cm}{\textbf{Monitoring and Updates}} & Model Performance Monitoring & Ongoing evaluation to ensure model performance doesn't degrade. & \$50,000 - \$200,000/year \cite{team_understanding_2024}, \cite{samarpit_what_2023} \\ \cline{2-4}
 & Regular Model Retraining & Required for maintaining relevance in dynamic fields like finance or health. & \$500,000 - \$2 million/year \cite{poole_what_2024}, \cite{team_understanding_2024} \\ \hline

\multirow{2}{3cm}{\textbf{Deployment Costs}} & API Management & Hosting, scaling, and managing APIs for model access. & \$1,000 - \$10,000/month \cite{team_understanding_2024}, \cite{noauthor_what_nodate} \\ \cline{2-4}
 & Scaling Infrastructure & Load balancing and redundancy for handling real-time traffic and requests. & \$10,000 - \$100,000/month \cite{team_understanding_2024}, \cite{samarpit_what_2023} \\ \hline

\textbf{Token-Related Cost} & Usage Case-Specific (Min - Max) & Token charges vary by use case (OpenAI GPT-3 costs  \$0.0001 - \$0.06 per token; Amazon Nova Pro \$0.80 - \$3.20 per token). & \$0.0001 - \$0.06/token or \$0.80 - \$3.20/token \cite{team_understanding_2024}, \cite{noauthor_how_nodate, amazon2024novareel} \\ \hline

\textbf{User Feedback Loop} & Continuous Improvement & Collecting user feedback to enhance model performance and personalization. & \$50,000 - \$200,000/year \cite{team_understanding_2024}, \cite{noauthor_what_nodate} \\ \hline

\textbf{Edge or On-Prem Deployment} & Specialized Hardware & On-prem hardware setups for privacy-focused or latency-sensitive tasks (e.g., NVIDIA A100). & \$100,000+ for setup; \$5,000 - \$50,000/month \cite{samarpit_what_2023}, \cite{poole_what_2024} \\ \hline

\end{tabular}
\end{table*}

\subsection{Environmental Concerns}

\begin{figure}
    \centering
    \includegraphics[width=1\linewidth]{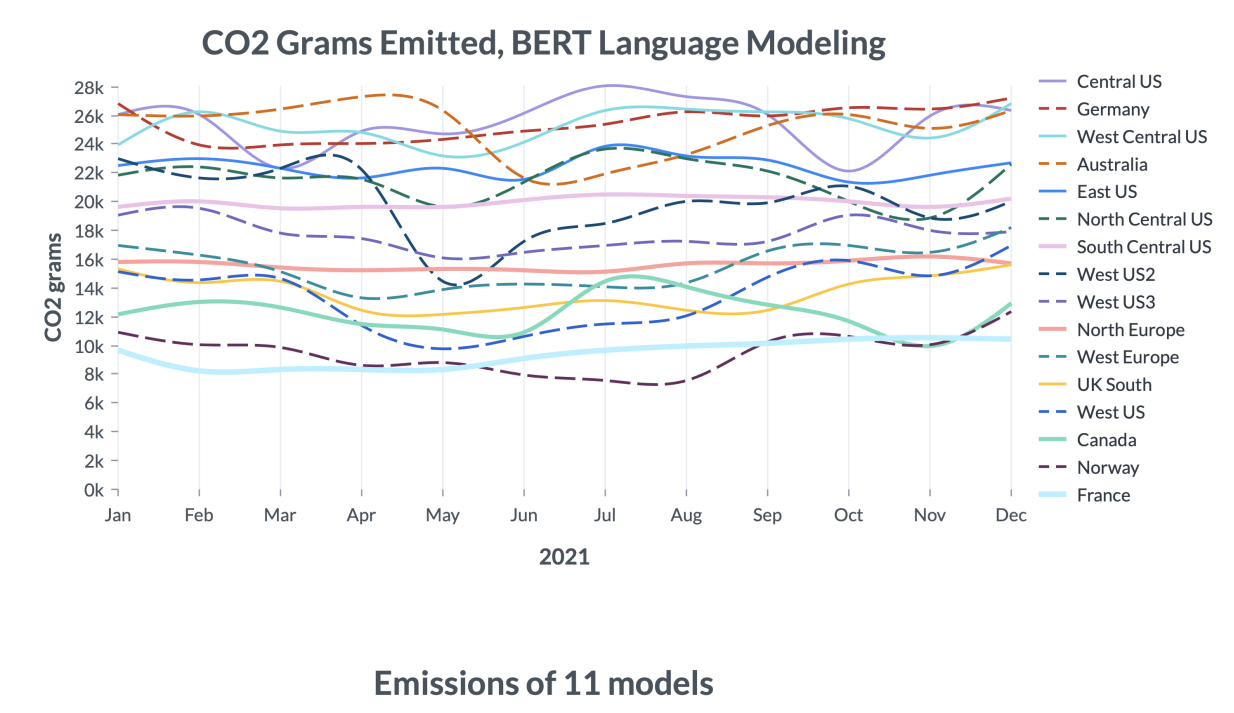}
    \caption{Carbon emissions for training BERT (8x V100s, 36 hrs): highest in Central US/Australia, lowest in Norway/France. From Dodge et al. \cite{dodge2022measuringcarbonintensityai}}
    \label{fig:CARBON11}
\end{figure}

\begin{figure}
    \centering
    \includegraphics[width=0.8\linewidth]{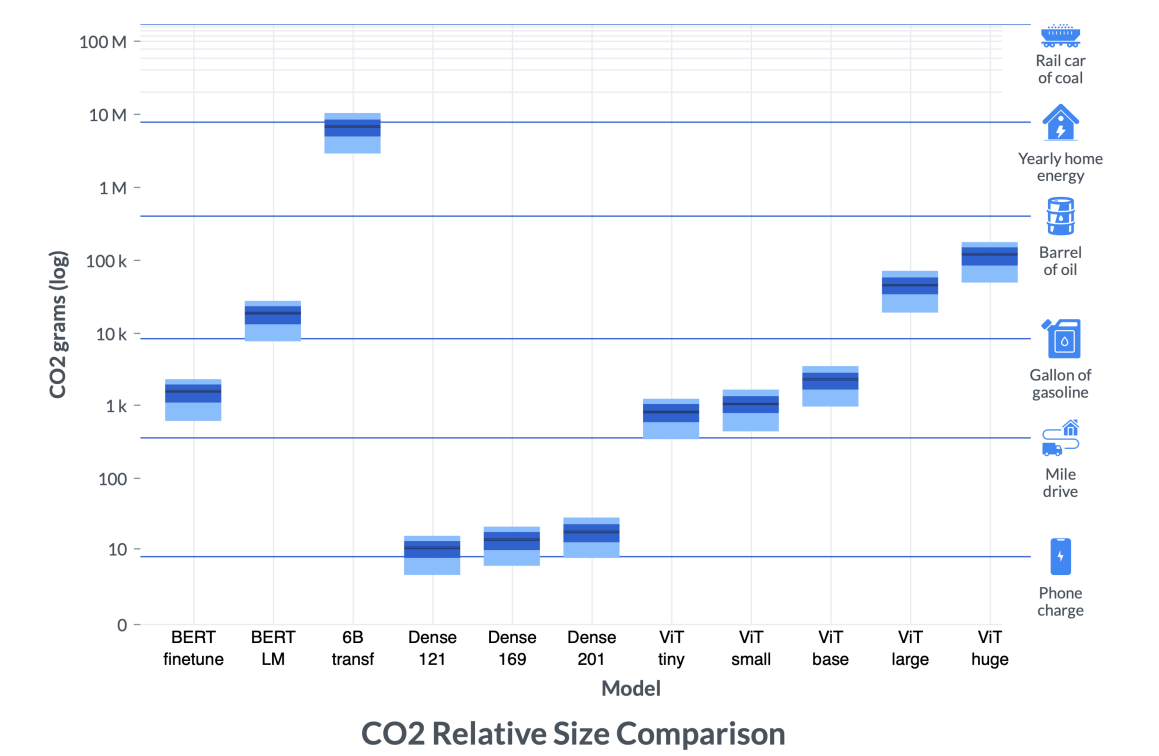}
    \caption{Carbon emissions for training experiments: larger models emit significantly more than smaller ones. From Dodge et al.\cite{dodge2022measuringcarbonintensityai}}
    \label{fig:CARBON}
\end{figure}

\begin{figure}[h] % Use [!t] to place the figure at the top of the page if possible
    \centering
    \includegraphics[width=0.4\textwidth]{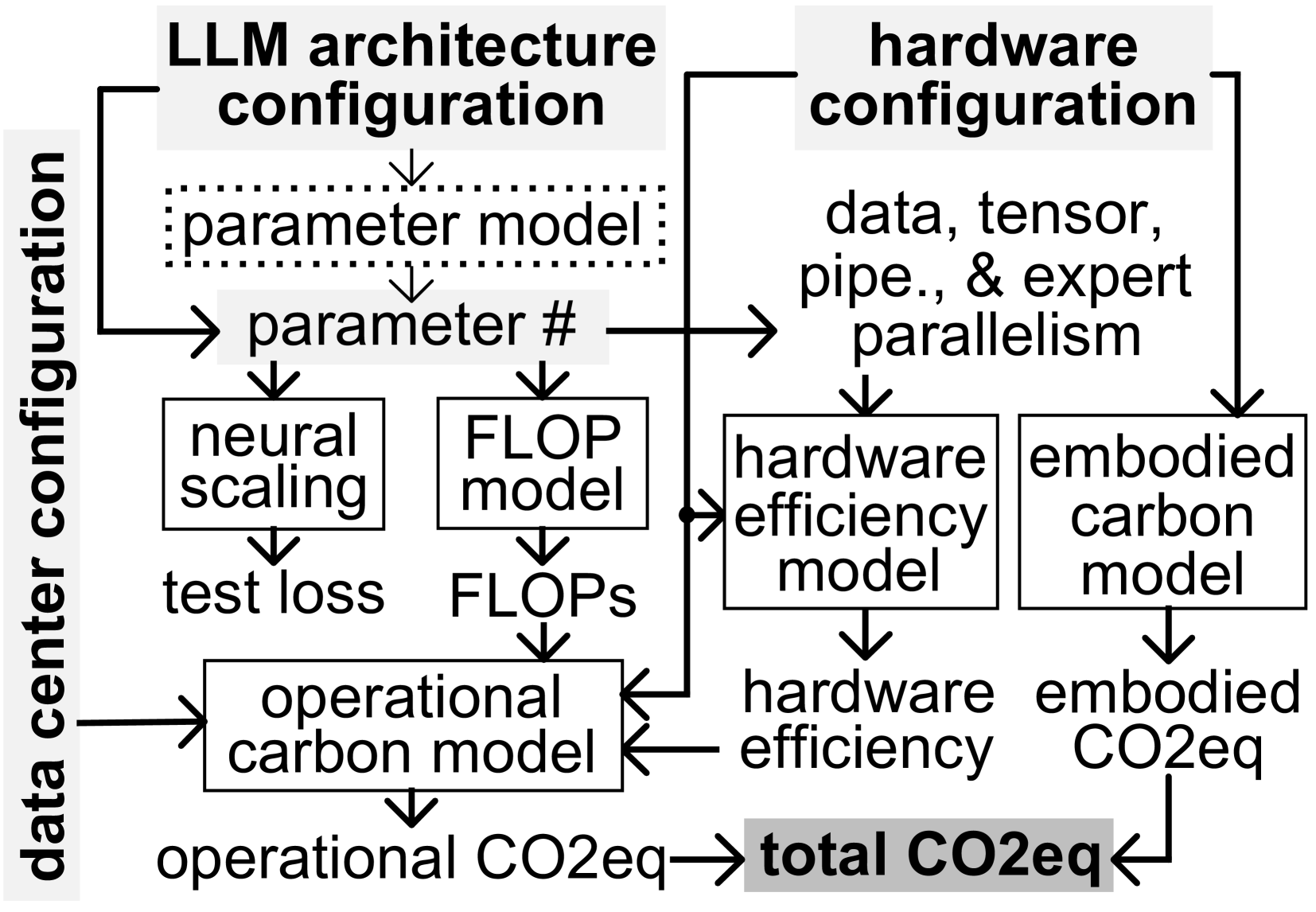} % Adjust the width to fit both columns
    \caption{LLMCarbon Overview \cite{faiz2024}}
    \label{fig:LLMCarbon Overview}
\end{figure}

As the adoption of LLMs grows, it is vital to address the environmental concerns associated with their development and deployment.  The environmental impact of Large Language Models (LLMs) is a growing concern, encompassing energy consumption, carbon emissions, and water usage. Below are key studies addressing these issues:

\begin{itemize}
    \item \textbf{Energy Consumption}: Training and operating Large Language Models (LLMs) require substantial computational resources, resulting in significant energy consumption. Faiz et al. \cite{faiz2024} introduce LLMCarbon, a framework to estimate energy usage throughout the LLM lifecycle, providing insights into the costs of training and inference. Similarly, \cite{SAMSI2023} analyzes the exponential growth in energy demands as model sizes and datasets expand, emphasizing the need for energy-efficient practices such as model pruning, quantization, and renewable energy integration. 
    Training phases are particularly energy-intensive, often requiring weeks of computation on large-scale GPU or TPU clusters, whereas inference, though less energy-demanding per task, becomes significant in high-frequency applications. Addressing these challenges requires optimizing both phases through advanced techniques like distributed learning and hardware improvements. These approaches are crucial for making LLMs more sustainable amidst growing computational demands.
    
    \item \textbf{Carbon Emissions}: The energy-intensive processes involved in Large Language Models (LLMs) contribute significantly to carbon dioxide equivalent (CO$_2$e) emissions. Carbon emissions vary significantly by model size and training location, as shown in Figures \ref{fig:CARBON11} and \ref{fig:CARBON}. The article "Double Jeopardy and Climate Impact in the Use of Large Language Models" highlights the carbon footprint arising from the training and operation of LLMs, emphasizing the importance of adopting sustainable AI practices \cite{solatorio2024doublejeopardyclimateimpact}. Faiz et al. \cite{faiz2024} introduced a framework, LLMCarbon (Fig. \ref{fig:LLMCarbon Overview}), to estimate emissions across the lifecycle of LLMs, including both operational and embodied carbon. Similarly, Luccioni et al. \cite{luccioni2022bloom} analyzed the BLOOM model’s emissions (Table \ref{tab:co2_emissions}), highlighting the statistics for equipement manufacturing, dynamic power consumption and idle energy consumption. To mitigate these emissions, strategies such as using energy-efficient hardware, implementing optimization techniques like pruning and quantization \cite{fu2023llmco2}, and transitioning to renewable energy sources \cite{deng2023autopcf} are recommended. These approaches are critical for reducing LLMs’ environmental impact while enabling sustainable AI development.    

    \item \textbf{Water Usage}: Data centers hosting LLMs require extensive cooling systems, leading to significant water consumption. Studies highlight the water usage associated with cooling data centers that support LLM infrastructure, emphasizing the environmental implications of such operations \cite{rillig2023risks}. Additional research emphasizes the substantial water consumption involved in training large AI models, advocating for transparency and strategies to mitigate this hidden environmental cost \cite{li2023thirsty}. The increasing demand for AI is leading to a surge in water usage by data centers, raising concerns about sustainability and resource management \cite{techcrunch2024water}. Furthermore, cooling methods employed in data centers and their associated water requirements highlight the challenges in balancing operational efficiency with environmental responsibility \cite{asce2024cooling}.

\end{itemize}

\begin{table}[htbp]
    \centering
    \caption{CO$_2$ Emissions Breakdown for BLOOM Model \cite{luccioni2022bloom}}
    \label{tab:co2_emissions}
    \begin{tabular}{|l|c|c|}
    \hline
    \textbf{Process}               & \textbf{CO$_2$ Emissions (tonnes)} & \textbf{Percentage} \\ \hline
    Equipment Manufacturing        & 11.2                              & 22.2\%              \\ \hline
    Dynamic Energy Consumption     & 24.69                             & 48.9\%              \\ \hline
    Idle Energy Consumption        & 14.6                              & 28.9\%              \\ \hline
    \end{tabular}
    \end{table}

\section{Sustainable Green AI}

The increasing demand for advanced artificial intelligence(AI) solutions necessitates the adoption of sustainable practices in AI development. This section explores methodologies aimed at minimizing energy consumption, optimizing resource utilization, and promoting environmentally friendly strategies throughout the AI lifecycle.

\subsection{Sustainable Practices}

Implementing sustainable practices in AI involves several key strategies:

\begin{itemize} \item \textbf{Energy-Efficient Training}: \begin{itemize} \item \textbf{Model Optimization Techniques}: Employing methods such as model pruning, quantization, and knowledge distillation can significantly reduce computational overhead during training, leading to lower energy consumption \cite{kim2021pqk}. \item \textbf{Sparsity-Focused Architectures}: Architectures like sparse transformers helps decrease energy usage without compromising model performance \cite{kim2021pqk}. \item \textbf{Distributed and Federated Learning}: Implementing distributed and federated learning frameworks decentralizes computational loads and uses localized data processing, enhancing energy efficiency \cite{kim2021pqk}. \end{itemize} \item \textbf{Sustainable Hardware}: \begin{itemize} \item \textbf{AI-Optimized Chips}: Leveraging specialized hardware such as Tensor Processing Units (TPUs) and low-power Graphics Processing Units (GPUs) designed for efficient computations can reduce energy consumption \cite{cao2024carbon}. \item \textbf{Neuromorphic Computing}: Exploring neuromorphic computing offers an alternative approach for AI tasks, potentially requiring significantly less energy \cite{neuromorphic}. \item \textbf{Edge AI Deployment}: Deploying AI computations on edge devices reduces dependence on power-intensive cloud servers, thereby conserving energy \cite{cao2024carbon}. \end{itemize} \item \textbf{Eco-Friendly Data Centers}: \begin{itemize} \item \textbf{Renewable Energy Integration}: Transitioning data centers to renewable energy sources, such as solar or wind power, minimizes carbon emissions \cite{fraunhofer}. \item \textbf{Innovative Cooling Systems}: Implementing advanced cooling solutions, including liquid immersion cooling, can significantly reduce energy consumption in data centers \cite{Neurram}. \item \textbf{Resource Virtualization}: Optimizing workloads through resource virtualization ensures maximum utilization of computing resources, enhancing energy efficiency \cite{fraunhofer}. \end{itemize} \end{itemize}

\subsection{Lifecycle Assessments}

Evaluating the complete lifecycle of AI models—from training and deployment to retirement—is crucial for ensuring sustainability at every stage:

\begin{itemize} \item \textbf{Lifecycle Optimization}: \begin{itemize} \item \textbf{Carbon Audits}: Conducting lifecycle carbon audits helps quantify and mitigate the environmental footprint of AI models \cite{neuromorphic}. \item \textbf{Model Reusability}: Enhancing model reusability by promoting shared repositories of pre-trained models for diverse tasks reduces the need for redundant training \cite{neuromorphic}. \item \textbf{Modular Design}: Designing AI systems with modular components allows for partial upgrades instead of full retraining, conserving resources \cite{neuromorphic}. \end{itemize} \item \textbf{Sustainable Deployment}: \begin{itemize} \item \textbf{Inference Efficiency}: Balancing inference efficiency with performance by optimizing real-time computation requirements reduces energy usage \cite{kim2021pqk}. \item \textbf{Adaptive Scaling}: Encouraging batch processing and adaptive scaling for AI services minimizes idle power consumption \cite{kim2021pqk}. \item \textbf{Low-Energy Solutions}: Using low-energy cloud services or on-premises solutions powered by renewable energy supports sustainable deployment \cite{fraunhofer}. \end{itemize} \item \textbf{End-of-Life Management}: \begin{itemize} \item \textbf{Decommissioning Protocols}: Developing protocols for responsibly decommissioning outdated AI systems ensures proper disposal and recycling \cite{Intel}. \item \textbf{Hardware Recycling}: Recycling or repurposing hardware used in AI model training and deployment reduces electronic waste \cite{Intel}. \item \textbf{Continuous Learning Frameworks}: Promoting continuous learning frameworks extends the usability of existing models without the need for retraining from scratch \cite{Intel}. \end{itemize} \end{itemize}

Adopting these sustainable practices in AI development can significantly reduce the environmental impact of technology while maintaining its innovative potential. These approaches ensure that advancements in AI align with global goals for environmental responsibility and sustainable growth.

\subsection{Case Study: Be.Ta Labs initiative towards sustainable AI}

The growth of AI technologies brings significant environmental challenges, with AI projected to contribute up to 5\% of global greenhouse gas emissions by 2030 \cite{andrae2020}. Be.Ta Labs is tackling this issue by leading the green AI movement, powering its entire AI and large language model (LLM) infrastructure with 100\% solar energy \cite{betalabs}. Through energy audits and the installation of a solar array, the company has cut its CO$_2$ emissions by over 90\% \cite{betalabsnews}.

In addition to using renewable energy, Be.Ta Labs employs proprietary energy-efficient techniques, enabling the training of a 70-billion-parameter model in just 1.2 kWh over 20 hours—dramatically reducing energy consumption compared to traditional methods \cite{betalabsnews}. These innovations lower both environmental and operational costs, making sustainable AI development more accessible.

Be.Ta Labs’ upcoming \textit{Aegis} project, the first LLM trained entirely on renewable energy, exemplifies their commitment to eco-friendly AI \cite{betalabs}. Their approach demonstrates that advanced AI can be developed sustainably, inspiring the tech industry to adopt greener practices and set new benchmarks for sustainability in technology.

\section{Conclusion}
This paper surveys the sustainability challenges and opportunities associated with Large Language Models (LLMs), focusing on their applications, economic impacts, and environmental costs. Key areas such as training methodologies, deployment strategies, and resource optimization are examined, leading to the identification of actionable insights to enhance the efficiency and sustainability of LLMs. Emphasis is placed on energy-efficient practices, renewable energy adoption, and lifecycle assessments to mitigate environmental impact. Solar-powered AI systems, such as those implemented by Be.Ta Labs, showcase the potential for significant reductions in carbon emissions, offering a scalable pathway for eco-friendly AI development. Looking forward, advancing solar-based infrastructure, enhancing energy-efficient hardware, and adopting modular AI architectures will be crucial in aligning the growth of LLMs with global sustainability goals.

\bibliographystyle{IEEEtran}
\bibliography{ref}

\end{document}